\documentclass[11pt,a4paper]{article}
\usepackage[hyperref]{acl2017}
\usepackage{times}
\usepackage{latexsym}
\usepackage{here}
\usepackage{bm}
\usepackage{url}
\usepackage{subcaption}
\usepackage{tabularx}
\usepackage{amsmath,amssymb}
\usepackage[pdftex]{graphicx}
\usepackage{CJKutf8}
\aclfinalcopy

\title{English-Japanese Neural Machine Translation with Encder-Decoder-Reconstructor}

\author{Yukio Matsumura \\
   \\
 \\\And
  Takayuki Sato \\
  Tokyo Metropolitan University \\
  Tokyo, Japan \\
   {\tt matsumura-yukio@ed.tmu.ac.jp, sasatatata99@gmail.com, komachi@tmu.ac.jp}  \\\And
  Mamoru Komachi \\
  \\
 \\}

\date{}

\begin{document}
\maketitle
\begin{abstract}
  Neural machine translation (NMT) has recently become popular in the field of machine translation. 
However, NMT suffers from the problem of repeating or missing words in the translation.
To address this problem, \newcite{tu2016neural} proposed an encoder-decoder-reconstructor framework for NMT using back-translation.
In this method, they selected the best forward translation model in the same manner as \newcite{DzmitryBahdana2014}, and then trained a bi-directional translation model as fine-tuning.
Their experiments show that it offers significant improvement in BLEU scores in Chinese-English translation task.
We confirm that our re-implementation also shows the same tendency and alleviates the problem of repeating and missing words in the translation on a English-Japanese task too.
In addition, we evaluate the effectiveness of pre-training by comparing it with a jointly-trained model of forward translation and back-translation.
\end{abstract}

\section{Introduction}

Recently, neural machine translation (NMT) has gained popularity in the field of machine translation. 
The conventional encoder-decoder NMT proposed by \newcite{Cho2014} uses two recurrent neural networks (RNN): one is an encoder, which encodes a source sequence into a fixed-length vector, and the other is a decoder, which decodes the vector into a target sequence.
A newly proposed attention-based NMT by \newcite{DzmitryBahdana2014} can predict output words using the weights of each hidden state of the encoder by the attention mechanism, improving the adequacy of translation.

Even with the success of attention-based models, a number of open questions remain in NMT.
\newcite{Tu2016} argued two of the common problems are over-translation: some words are repeatedly translated unnecessary and under-translation: some words are mistakenly untranslated.
This is due to the fact that NMT can not completely convert the information from the source sentence to the target sentence.
\newcite{Mi2016a} and \newcite{Feng2016} pointed out that NMT lacks the notion of coverage vector in phrase-based statistical machine translation (PBSMT), so unless otherwise specified, there is no way to prevent missing translations. 

Another problem in NMT is an objective function.
NMT is optimized by cross-entropy; therefore, it does not directly maximize the translation accuracy. 
\newcite{Shen2016} pointed out that optimization by cross-entropy is not appropriate and proposed a method of optimization based on a translation accuracy score, such as expected BLEU, which led to improvement of translation accuracy. 
However, BLEU is an evaluation metric based on n-gram precision; therefore, repetition of some words may be present in the translation even though the BLEU score is improved. 

\begin{figure*}[t]
\begin{minipage}{0.5\hsize}
\centering
\includegraphics[height=4.5cm]{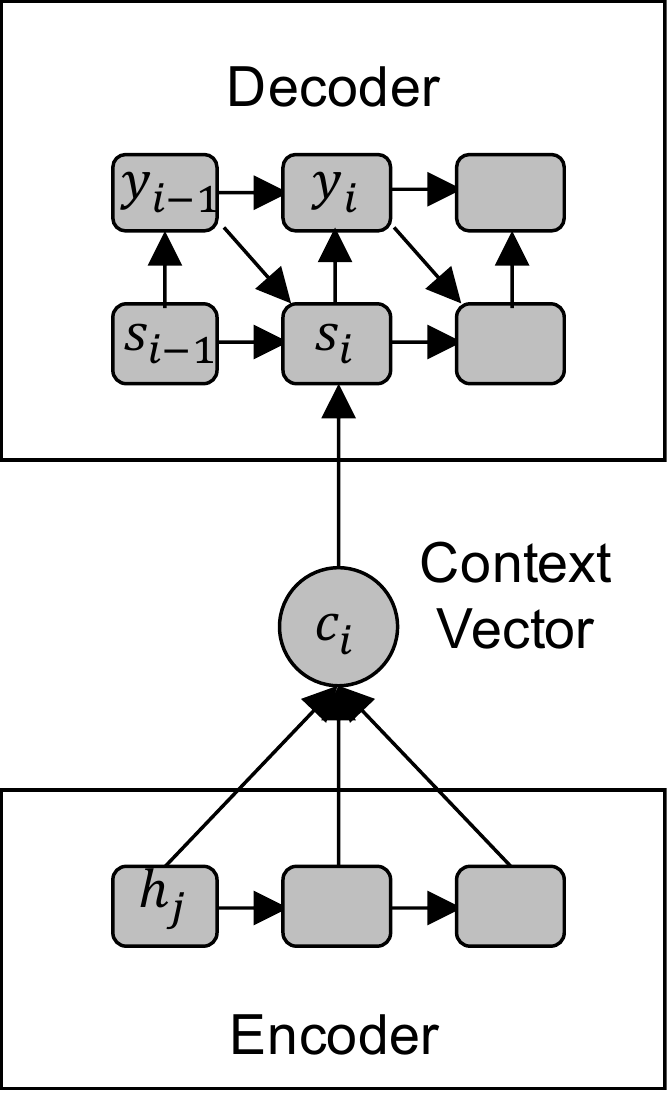}
\caption{Attention-based NMT.}
\label{FNMT}
\end{minipage}
\begin{minipage}{0.5\hsize}
\centering
\includegraphics[height=4.5cm]{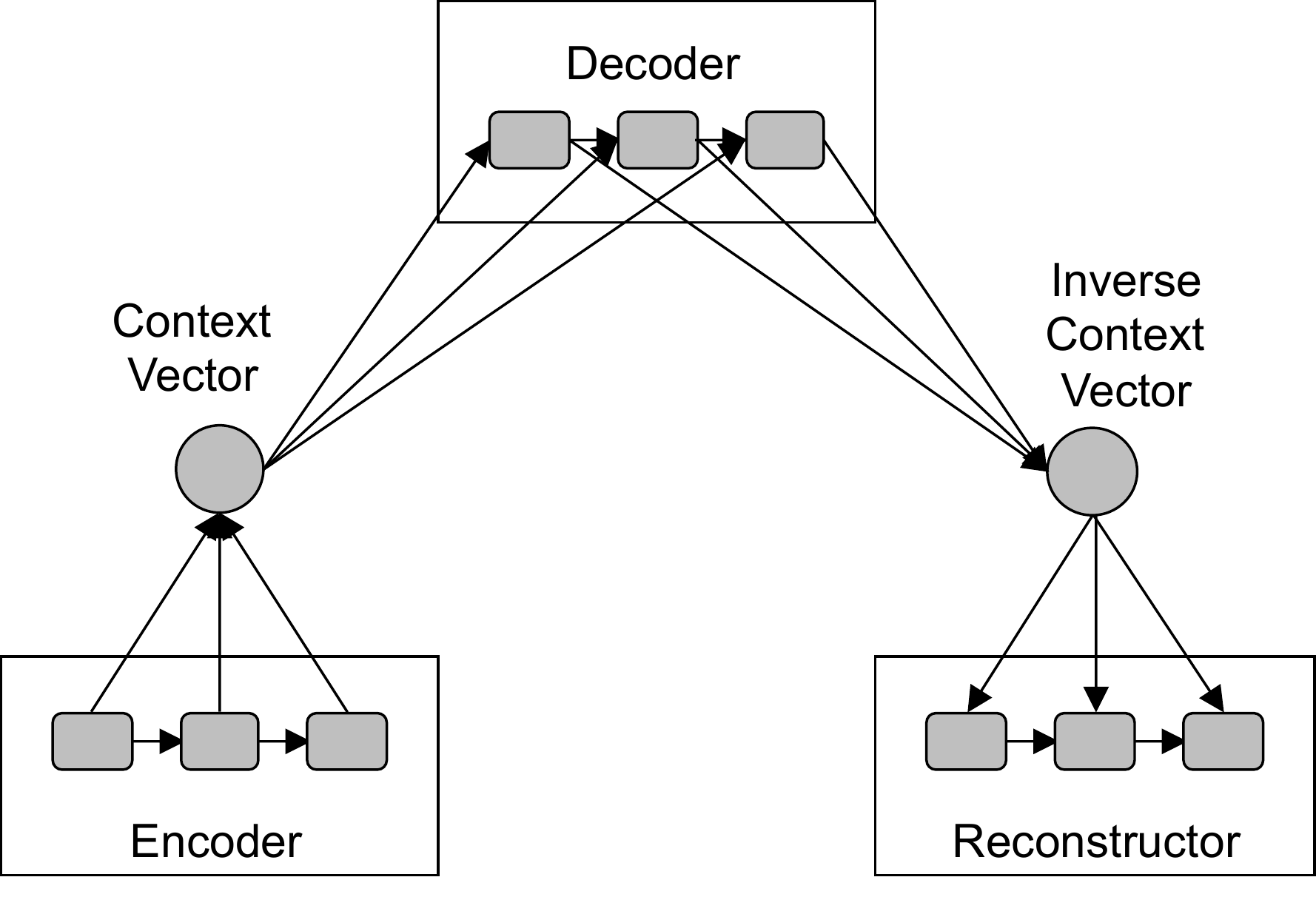}
\caption{Encoder-Decoder-Reconstructor.}
\label{FBTONMT}
\end{minipage}
\end{figure*}

To address to problem of repeating and missing words in the translation, \newcite{tu2016neural} introduce an encoder-decoder-reconstructor framework that optimizes NMT by back-translation from the output sentences into the original source sentences.
In their method, after training the forward translation in a manner similar to the conventional attention-based NMT, they train a back-translation model from the hidden state of the decoder into the source sequence by a new decoder to enforce agreement between source and target sentences.

In order to confirm the language independence of the framework, we experiment on two parallel corpora of English-Japanese and Japanese-English translation tasks using encode-decoder-reconstructor.
Our experiments show that their method offers significant improvement in BLEU scores and alleviates the problem of repeating and missing words in the translation on English-Japanese translation task, though the difference is not significant on Japanese-English translation task.

In addition, we jointly train a model of forward translation and back-translation without pre-training, and then evaluate this model.
As a result, the encoder-decoder-reconstructor can not be trained well without pre-training, so it proves that we have to train the forward translation model in a manner similar to the conventional attention-based NMT as pre-training.

The main contributions of this paper are as follows:
\begin{itemize}
\item Experimental results show that encode-decoder-reconstructor framework achieves significant improvements in BLEU scores (1.0-1.4) for English-Japanese translation task.
\item Experimental results show that encode-decoder-reconstructor framework has to train the forward translation model in a manner similar to the conventional attention-based NMT as pre-training.
\end{itemize}

\section{Related Works}
Several studies have addressed the NMT-specific problem of missing or repeating words.
\newcite{Niehues2016} optimized NMT by adding the outputs of PBSMT to the input of NMT.
\newcite{Mi2016a} and \newcite{Feng2016} introduced a distributed version of coverage vector taken from PBSMT to consider which words have been already translated.
All these methods, including ours, employ information of the source sentence to improve the quality of translation, but our method uses back-translation to ensure that there is no inconsistency.
Unlike other methods, once learned, our method is identical to the conventional NMT model, so it does not need any additional parameters such as coverage vector or a PBSMT system for testing. 

The attention mechanism proposed by \newcite{Meng2016} considers not only the hidden states of the encoder but also the hidden states of the decoder so that over-translation can be relaxed.
In addition, the attention mechanism proposed by \newcite{Feng2016} computes a context vector by considering the previous context vector to prevent over-translation.
These works indirectly reduce repeating and missing words, while we directly penalize translation mismatch by considering back-translation. 

The encoder-decoder-reconstructor framework for NMT proposed by \newcite{tu2016neural} optimizes NMT by reconstructor using back-translation. 
They consider likelihood of both of forward translation and back-translation, and then this framework offers significant improvement in BLEU scores and alleviates the problem of repeating and missing words in the translation on a Chinese-English translation task.

\section{Neural Machine Translation}

Here, we describe the attention-based NMT proposed by \newcite{DzmitryBahdana2014} as shown in Figure \ref{FNMT}. 

The input sequence ($\bm{x}=[x_1, x_2, \cdots, x_{|\bm{x}|}]$) is converted into a fixed-length vector by the encoder using an RNN.
At each time step $t$, the hidden state $h_t$ of the encoder is presented as 
\begin{equation}
h_t = [\overrightarrow{h_t}^{\top}:\overleftarrow{h_t}^{\top}]^{\top}
\end{equation}
using a bidirectional RNN. 
The forward state $\overrightarrow{h_t}$ and the backward state $\overleftarrow{h_t}$ are computed by
\begin{equation}
\overrightarrow{h_t} = r(x_t, h_{t-1})
\end{equation}
and
\begin{equation}
\overleftarrow{h_t} = r'(x_t, h_{t+1})
\end{equation}
where $r$ and $r'$ are nonlinear functions. 
The hidden states $(h_1, h_2, \cdots, h_{|\bm{x}|})$ are converted into a fixed-length vector $v$ as
\begin{equation}
v = q([h_1, h_2, \cdots, h_{|\bm{x}|}])
\end{equation}
where $q$ is a nonlinear function.  

The fixed-length vector $v$ generated by the encoder is converted into the target sequence ($\bm{y}=[y_1, y_2, \cdots, y_{|\bm{y}|}]$) by the decoder using an RNN.
At each time step $i$, the conditional probability of the output word $\hat{y}_i$  is computed by
\begin{equation}
p(\hat{y}_i|\bm{y}_{<i}, \bm{x}) = f(s_i, y_{i-1}, c_i)
\end{equation}
where $f$ is a nonlinear function. 
The hidden state $s_i$ of the decoder is presented as
\begin{equation}
s_i = g(s_{i-1}, y_{i-1}, c_i)
\end{equation}
using the hidden state $s_{i-1}$ and the target word $y_{i-1}$ at the previous time step and the context vector $c_i$.  

The context vector $c_i$ is a weighted sum of each hidden state $h_j$ of the encoder. 
It is presented as 
\begin{equation}
c_i=\sum_{j=1}^{|x|} \alpha_{ij}h_j
\end{equation}
and its weight $\alpha_{ij}$ is a normalized probability distribution.
It is computed by 
\begin{equation}
\alpha_{ij}=\frac{\exp(e_{ij})}{\sum_{k=1}^{|x|} \exp(e_{ik})}
\end{equation}
and
\begin{equation}
e_{ij}=v_a^{\top}\tanh(W_as_{i-1} + U_ah_j)
\end{equation}
where $v_a$ is a weight vector and $W_a$ and $U_a$ are weight matrices.  

The objective function is defined by  
\begin{equation}
\label{Ldef}
\begin{split}
\mathcal{L}(\theta)&=\frac{1}{N}\sum\nolimits_{n=1}^{N} \sum_{i=1}^{|y|}\log{p(\hat{y}_i^{(n)}|\bm{y}_{<i}^{(n)}, \bm{x}^{(n)}, \theta)} 
\end{split}
\end{equation}
where $N$ is the number of data and $\theta$ is a model parameter. 

Incidentally, as a nonlinear function, the hyperbolic tangent function or the rectified linear unit are generally used. 

\begin{table}[t]
\centering
\begin{tabular}{c|rr} \hline
\multicolumn{1}{c|}{} & \multicolumn{1}{c}{ASPEC} & \multicolumn{1}{c}{NTCIR}\\ \hline
train & 827,188 & 1,169,201 \\ 
dev & 1,504 & 2,741 \\ 
test & 1,556 & 2,300 \\ \hline
\end{tabular}
\vspace{-5pt}
\caption{Numbers of parallel sentences.}
\label{corpus}
\end{table}

 \begin{table*}[t]
\centering
\begin{tabular}{c|l|ccc} \hline
\multicolumn{5}{c}{English-Japanese} \\ \hline
\multicolumn{1}{c|}{Corpus} &\multicolumn{1}{c|}{Model} & \multicolumn{1}{c}{BLEU} & \multicolumn{1}{c}{$p$-value} & \multicolumn{1}{c}{Hours} \\ \hline
 & Baseline-NMT & 29.75 & - & 99\\ 
ASPEC & +Reconstructor & 30.76  & 0.00 & 149\\ 
  & +Reconstructor (Jointly-Training) & 26.04 & - & 174\\ \hline
 & Baseline-NMT &30.03 & - & 116\\ 
NTCIR & +Reconstructor & 31.40 & 0.00 & 166\\ 
  & +Reconstructor (Jointly-Training)  & 29.04 & - & 252\\ \hline
\end{tabular}
\vspace{-5pt}
\caption{English-Japanese translation results.}
 \label{resultEJ}
\vspace{10pt}
\end{table*}

 \begin{table*}[t]
\centering
\begin{tabular}{c|l|ccc} \hline
\multicolumn{5}{c}{Japanese-English}\\ \hline
\multicolumn{1}{c|}{Corpus} &\multicolumn{1}{c|}{Model} & \multicolumn{1}{c}{BLEU} & \multicolumn{1}{c}{$p$-value} & \multicolumn{1}{c}{Hours} \\ \hline
 & Baseline-NMT & 21.91 & - & 87 \\ 
ASPEC & +Reconstructor & 22.27 & 0.10 & 127 \\ 
  & +Reconstructor (Jointly-Training) & 16.29 & - & 187 \\ \hline
 & Baseline-NMT & 29.48 & - & 180 \\ 
NTCIR & +Reconstructor & 29.73 & 0.11 & 244\\ 
  & +Reconstructor (Jointly-Training)  & 28.95 & - & 300\\ \hline
\end{tabular}
\vspace{-5pt}
\caption{Japanese-English translation results.}
 \label{resultJE}
\vspace{10pt}
\end{table*}

\section{Encoder-Decoder-Reconstructor}

\subsection{Architecture}
Next, we describe the encoder-decoder-reconstructor framework for NMT proposed by \newcite{tu2016neural} as shown in Figure \ref{FBTONMT}. 
The encoder-decoder-reconstructor consists of two components: the standard encoder-decoder as an attention-based NMT proposed by \newcite{DzmitryBahdana2014} and the \textbf{reconstructor} which back-translates from the hidden states of decoder to the source sentence.

In their method, the hidden state of the decoder is back-translated into the source sequence ($\bm{x}$) by the reconstructor for the back-translation.
At each time step $i$, the conditional probability of the output word $\hat{x}_i$  is computed by 
\begin{equation}
p(\hat{x}_i|\bm{x}_{<i}, \bm{\hat{y}}) = f'(s'_i, x_{i-1}, c'_i)
\end{equation}
where $f'$ is a nonlinear function. 
The hidden state $s'_i$ of the reconstructor is presented as
\begin{equation}
s'_i = g'(s'_{i-1}, x_{i-1}, c'_i)
\end{equation}
using the hidden state $s'_{i-1}$ and the source word $x_{i-1}$ at the previous time step and the new context vector (\textbf{inverse context vector}) $c'_i$.  

The inverse context vector $c'_i$ is a weighted sum of each hidden state $s_j$ of the decoder (on forward translation). 
It is presented as 
\begin{equation}
c'_i=\sum_{j=1}^{|y|} \alpha'_{ij}s_j
\end{equation}
and its weight $\alpha'_{ij}$ is a normalized probability distribution.
It is computed by 
\begin{equation}
\alpha'_{ij}=\frac{\exp(e'_{ij})}{\sum_{k=1}^{|y|} \exp(e'_{ik})}
\end{equation}
and
\begin{equation}
e'_{ij}=v'_a{\top}\tanh(W'_as'_{i-1} + U'_as_j)
\end{equation}
where $v'_a$ is a weight vector and $W'_a$ and $U'_a$ are weight matrices.  

The objective function is defined by   
\begin{equation}
\label{L}
\begin{split}
\mathcal{L}(\theta, \gamma)&=\frac{1}{N}\sum\nolimits_{n=1}^{N} \Bigl\{ \sum_{i=1}^{|y|}\log{p(\hat{y}_i^{(n)}|\bm{y}_{<i}^{(n)}, \bm{x}^{(n)}, \theta)} \\ 
&\quad +\lambda\sum_{i=1}^{|x|}\log{p(\hat{x}_i^{(n)}|\bm{x}_{<i}^{(n)}, \bm{s}^{(n)}, \gamma)} \Bigr\} 
\end{split}
\end{equation}
where $N$ is the number of data, $\theta$ and $\gamma$ are model parameters and $\lambda$ is a hyper-parameter which can consider the weight between forward translation and back-translation.

This objective function consists of two parts: forward measures translation fluency, and backward measures translation adequacy. Thus, the combined objective function is more consistent with the goal of enhancing overall translation quality, and can more effectively guide the parameter training for making better translation.

 \begin{table*}[t]
\centering
\begin{tabularx}{\linewidth}{c|X} \hline
\multicolumn{2}{c}{Example 1: Improvement in under-translation.} \\ \hline
Input & the conditions under which the effect of turbulent viscosity is correctly evaluated were examined \underline{\textit{on the basis of the relation between turbulent}} \underline{\textit{viscosity and numerical viscosity in size}} . \\ \hline
Baseline-NMT & \mbox{\begin{CJK}{UTF8}{min}乱 流 粘性 の 影響 を 正確 に 評価 する 条件 を 検討 し た 。\end{CJK}}\\ 
\, & \\ \hline
+Reconstructor &  \mbox{\begin{CJK}{UTF8}{min}乱 流 粘性 の 影響 を 正確 に 評価 する 条件 を , \underline{\textbf{乱 流 粘性 と 数値 的 粘性}}\end{CJK}}\\ 
\, & \mbox{\begin{CJK}{UTF8}{min}\underline{\textbf{の 関係 を 基 に}} 調べ た 。\end{CJK}} \\ \hline
+Reconstructor & \mbox{\begin{CJK}{UTF8}{min}乱 流 粘性 の 影響 を 考慮 し た 条件 を , \underline{\textbf{乱 流 粘性 と 粘性 の 粘性 と の}}\end{CJK}}\\ 
\,(Jointly-Training) & \mbox{\begin{CJK}{UTF8}{min}\underline{\textbf{関係 を もと に}} 検討 し た 。\end{CJK}}\\ \hline
Reference & \mbox{\begin{CJK}{UTF8}{min}\underline{\textbf{乱 流 粘性 と 数値 粘性 の 大小 関係 により , }}乱 流 粘性 の 効果 が 正しく\end{CJK}} \mbox{\begin{CJK}{UTF8}{min}評価 さ れる 条件 を 検討 し た 。\end{CJK}} \\ \hline
\end{tabularx} \\

\begin{tabularx}{\linewidth}{c|X} \hline
\multicolumn{2}{c}{Example 2: Improvement in over-translation.} \\ \hline
Input & activity was high in cells of the young , especially \underline{\textit{newborn infant}} , and was very slight in cells \underline{\textit{of 30 ‐ year ‐ old or more}} .\\ \hline
Baseline-NMT & \mbox{\begin{CJK}{UTF8}{min}活動 性 は 若 齢 , 特に \underline{\textbf{新生児 新生児}} で は \underline{\textbf{30 歳 以上 の}} 細胞 で 高く , \end{CJK}}\\ 
\, & \mbox{\begin{CJK}{UTF8}{min}\underline{\textbf{30 歳 以上 の}} 細胞 で は わずか で あっ た 。\end{CJK}}\\ \hline
+Reconstructor & \mbox{\begin{CJK}{UTF8}{min}その 活性 は 若 齢 , 特に \underline{\textbf{新生児}} は 細胞 が 高く , \underline{\textbf{30 歳 以上 の}} 細胞 で は\end{CJK}} \\
\, & \mbox{\begin{CJK}{UTF8}{min}わずか で あっ た 。\end{CJK}}\\ \hline
+Reconstructor & \mbox{\begin{CJK}{UTF8}{min}若 齢 の \underline{\textbf{新生児}} で は 活性 は 高かっ た が , \underline{\textbf{30 歳 以上 の}} 場合 に は 極めて\end{CJK}}\\
\,(Jointly-Training) & \mbox{\begin{CJK}{UTF8}{min}軽度 で あっ た 。\end{CJK}}\\ \hline
Reference & \mbox{\begin{CJK}{UTF8}{min}活性 は 若い 個体 , 特に \underline{\textbf{新生児}} の 細胞 で 高く , \underline{\textbf{30 歳 以上 の}} もの で は\end{CJK}} \mbox{\begin{CJK}{UTF8}{min}ごく わずか で あっ た 。 \end{CJK}}\\
\end{tabularx}
\vspace{-6pt}
\caption{Examples of outputs of English-Japanese translation.}
\label{sample}
\vspace{10pt}
\end{table*}

\begin{figure*}[t]
\vspace{-50pt}
\hspace{-11pt}
\begin{minipage}{0.4\hsize}
\centering
\includegraphics[height=8cm]{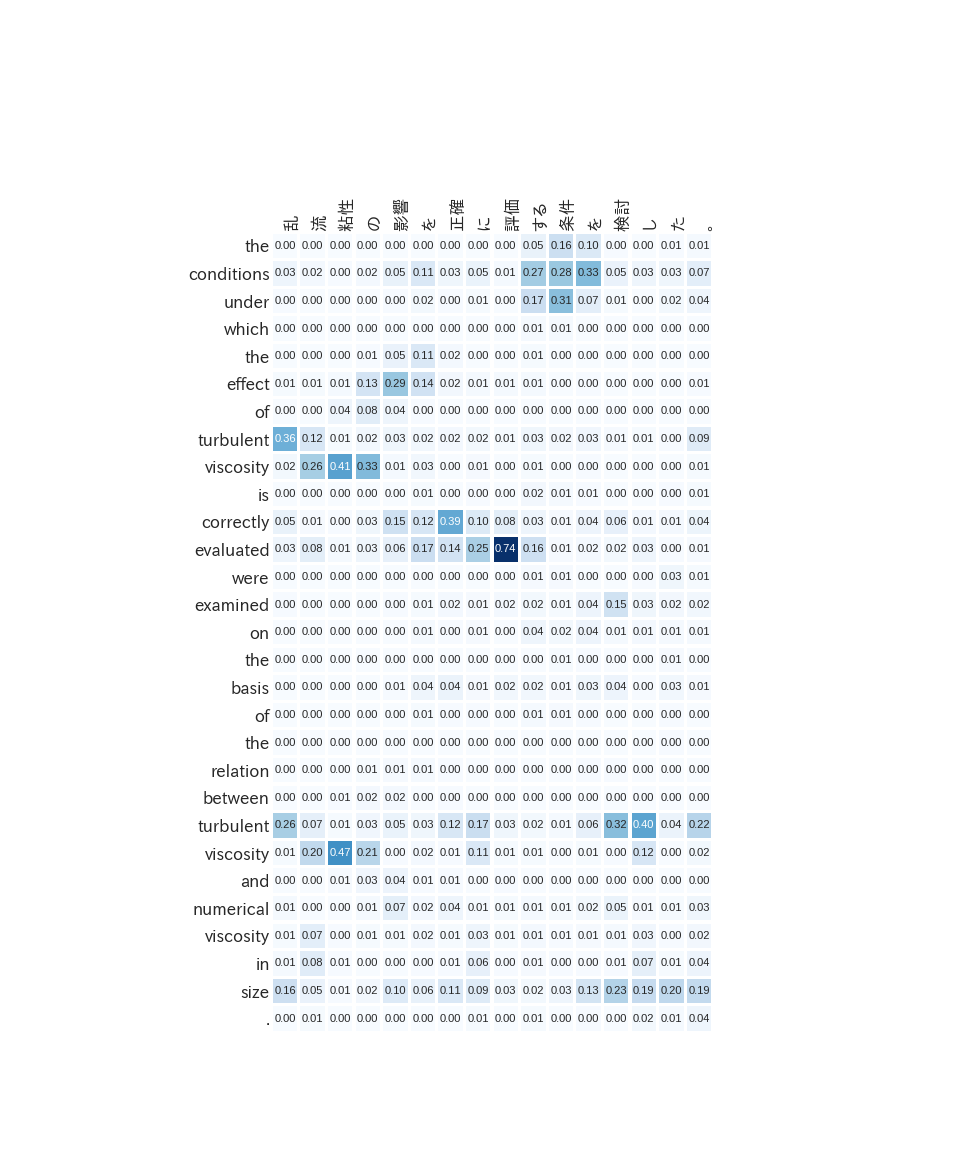}
\mbox{\hspace{11pt}Baseline-NMT}
\end{minipage}
\begin{minipage}{0.6\hsize}
\centering
\includegraphics[height=8cm]{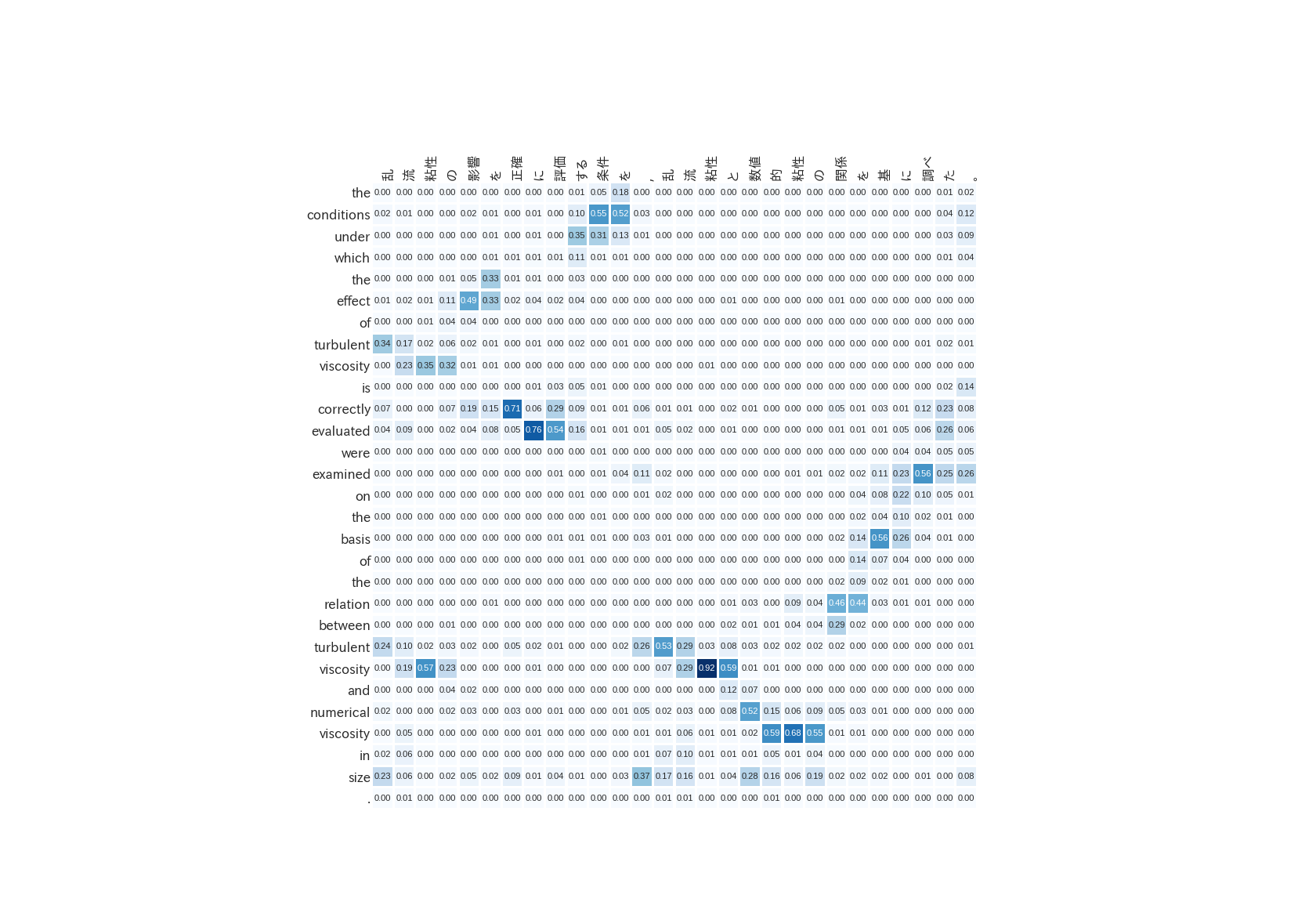}
\mbox{\hspace{65pt}Encoder-Decoder-Reconstructor}
\end{minipage}
\caption{The attention layer in Example 1 : Improvement in under-translation.}
\label{att1}

\hspace{-88pt}
\begin{minipage}{0.62\hsize}
\centering
\includegraphics[height=8cm]{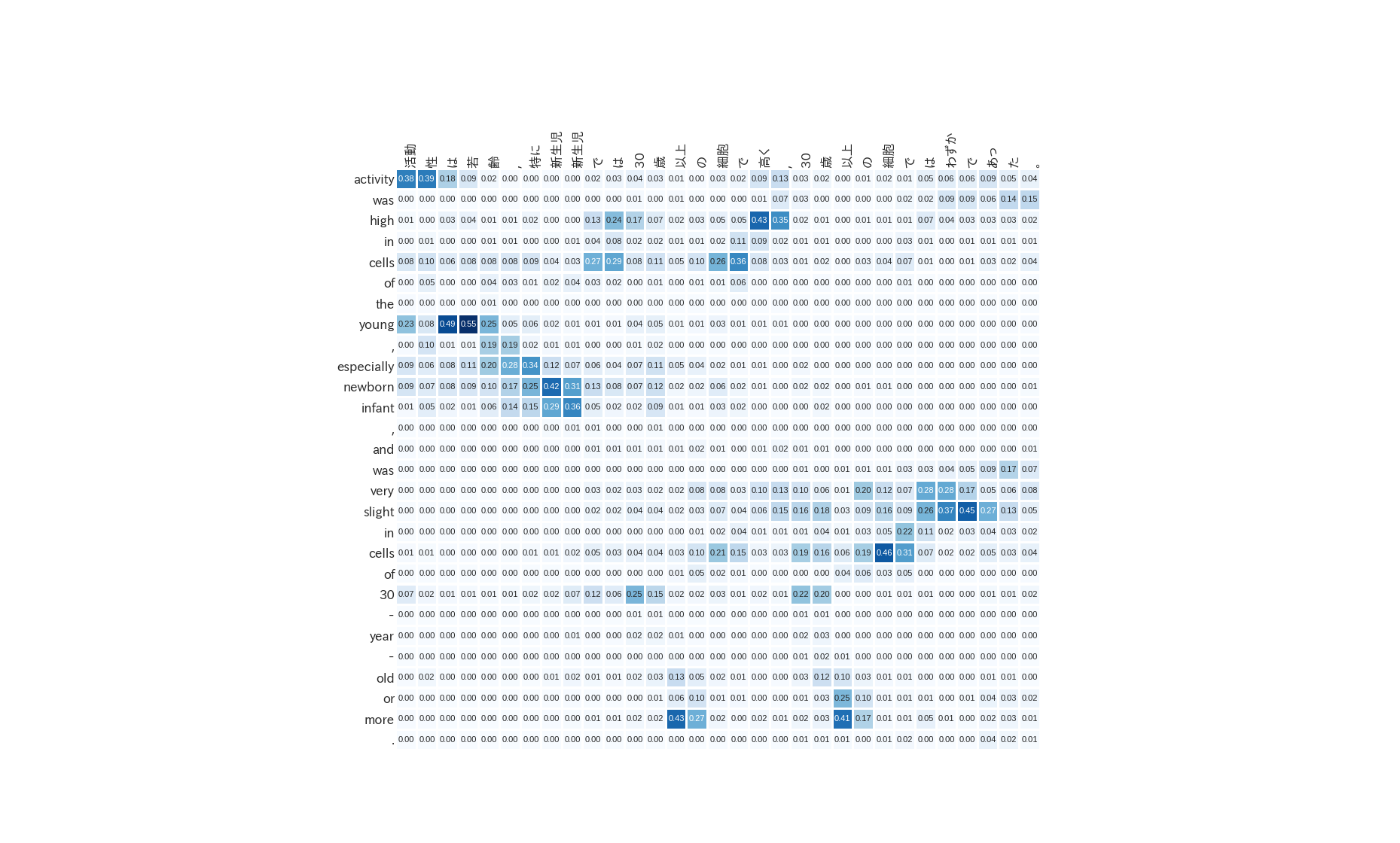}
\mbox{\hspace{97pt}Baseline-NMT}
\end{minipage}
\begin{minipage}{0.38\hsize}
\centering
\includegraphics[height=8cm]{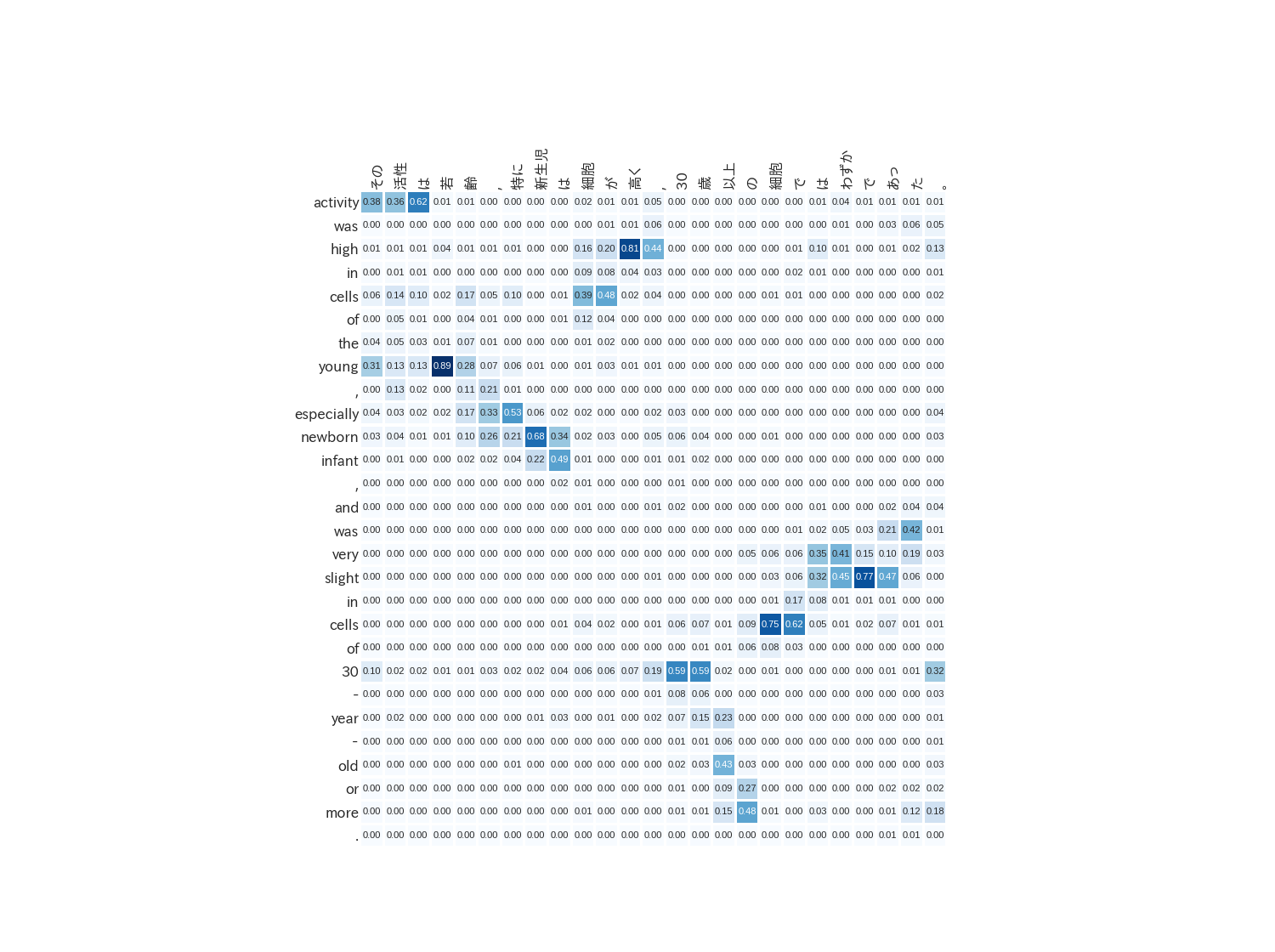}
\mbox{\hspace{85pt}Encoder-Decoder-Reconstructor}
\end{minipage}
\caption{The attention layer in Example 2 : Improvement in over-translation.}
\label{att2}
\end{figure*}

\subsection{Training}
The encoder-decoder-reconstructor is trained with likelihood of both the encoder-decoder and the reconstructor on a set of training datasets. 
\newcite{tu2016neural} trained a back-translation model from the hidden state of the decoder into the source sequence by reconstructor to enforce agreement between source and target sentences using Equation \ref{L} after training the forward translation in a manner similar to the conventional attention-based NMT using Equation \ref{Ldef}.

In addition, we experiment to jointly train a model of forward translation and back-translation without pre-training.
It may learn a globally optimal model compared to locally optimal model pre-trained using the forward translation.

\subsection{Testing}
\newcite{tu2016neural} used a beam search to predict target sentences that approximately maximizes both of forward translation and back-translation on testing.
In this paper, however, we do not use a beam search for simplicity and effectiveness. 

\begin{table*}
\centering
\begin{tabular}{c|l|rrr|rrr} \hline
\multicolumn{1}{c|}{Corpus} &\multicolumn{1}{c|}{Model} & \multicolumn{3}{c|}{English-Japanese} & \multicolumn{3}{c}{Japanese-English}\\ \cline{3-8}
\multicolumn{1}{c|}{} & \multicolumn{1}{c|}{} & \multicolumn{1}{c}{(\romannumeral1)} & \multicolumn{1}{c}{(\romannumeral2)} & \multicolumn{1}{c|}{(\romannumeral3)} & \multicolumn{1}{c}{(\romannumeral1)} & \multicolumn{1}{c}{(\romannumeral2)} & \multicolumn{1}{c}{(\romannumeral3)}\\ \hline
 & Baseline-NMT & 1,141 & 378 & 1,045 & 951 & 494 & 1,085 \\ 
ASPEC & +Reconstructor & 988 & 336 & 1,042 & 836 & 418 & 1,014 \\ 
 & +Reconstructor (Jointly-Training) & 1,292 & 446 & 1,147 & 1,106 & 525 & 1,821 \\ \hline
 & Baseline-NMT & 2,122 & 1,015 & 1,106 & 2,521 & 1,073 & 1,630\\ 
NTCIR & +Reconstructor & 1,958 & 922 & 963 & 2,187 & 987 & 1,422 \\ 
  & +Reconstructor (Jointly-Training) & 1,978 & 916 & 1,078 & 2,475 & 1,107 & 1,610 \\ \hline
\end{tabular}
\vspace{-5pt}
\caption{Numbers of redundant and unknown word tokens.}
 \label{repete}
\end{table*}

\section{Experiments}
We evaluated the encoder-decoder-reconstructor framework for NMT on English-Japanese and Japanese-English translation tasks.
\subsection{Datasets}
We used two parallel corpora: Asian Scientific Paper Excerpt Corpus (ASPEC) \cite{Nakazawa2016} and NTCIR PatentMT Parallel Corpus \cite{Goto2013}. 
Regarding the training data of ASPEC,  we used only the first 1 million sentences sorted by sentence-alignment similarity.
Japanese sentences were segmented by the morphological analyzer MeCab (version 0.996, IPADIC), and English sentences were tokenized by tokenizer.perl of Moses. 
Table \ref{corpus} shows the numbers of the sentences in each corpus.
Note that sentences with more than 40 words were excluded from the training data.

\subsection{Models}
We used the attention-based NMT \cite{DzmitryBahdana2014} as a baseline-NMT, the encoder-decoder-reconstructor \cite{tu2016neural} and the encoder-decoder-reconstructor that jointly trained forward translation and back-translation without pre-training.
The RNN used in the experiments had 512 hidden units, 512 embedding units, 30,000 vocabulary size and 64 batch size.
We used Adagrad (initial learning rate 0.01) for optimizing model parameters. 
We trained our model on GeForce GTX TITAN X GPU.
Note that we set the hyper-parameter $\lambda = 1$ on the encoder-decoder-reconstructor same as \newcite{tu2016neural}.

\subsection{Results}
Tables \ref{resultEJ} and \ref{resultJE} show the translation accuracy in BLEU scores, the $p$-value of the significance test by bootstrap resampling \cite{Koehn2004} and training time in hours until convergence.
The encoder-decoder-reconstructor \cite{tu2016neural} requires slightly longer time to train than the baseline NMT, but we emphasize that decoding time remains the same with the encoder-decoder-reconstructor and baseline-NMT.
The results show that the encoder-decoder-reconstructor \cite{tu2016neural} significantly improves translation accuracy by 1.01 points on ASPEC and 1.37 points on NTCIR in English-Japanese translation ($p < 0.05$). However, it does not significantly improve translation accuracy in Japanese-English translation.
In addition, it is proved that the encoder-decoder-reconstructor without pre-training worsens rather than improves translation accuracy.

Table \ref{sample} shows examples of outputs of English-Japanese translations.
In Example 1, ``\begin{CJK}{UTF8}{min}乱 流 粘性 と 数値 粘性 の 大小 関係 により ,\end{CJK}'' (on the basis of the relation between turbulent viscosity and numerical viscosity in size) is missing in the output of baseline-NMT, but ``\begin{CJK}{UTF8}{min}乱 流 粘性 と 数値 的 粘性 の 関係 を 基 に\end{CJK}'' (on the basis of the relation between turbulent viscosity and numerical viscosity) is present in the output of encoder-decoder-reconstructor.
In Example 2, ``\begin{CJK}{UTF8}{min}新生児\end{CJK}'' (newborn infant) and ``\begin{CJK}{UTF8}{min}30歳以上の\end{CJK}'' (of 30 ‐ year ‐ old or more) are repeated in the output of baseline-NMT, but they appear only once in the output of encoder-decoder-reconstructor. 

In addition, Figures \ref{att1} and \ref{att2} show the attention layer on baseline-NMT and encoder-decoder-reconstructor in each example. 
In Figure \ref{att1}, although the attention layer of baseline NMT attends input word ``turbulent'', the decoder does not output ``\begin{CJK}{UTF8}{min}乱流\end{CJK}'' (turbulent) but ``\begin{CJK}{UTF8}{min}検討\end{CJK}'' (examined) at the 13th word.
Thus, under-translation may be resulted from the hidden layer or the embedding layer instead of the attention layer.
In Figure \ref{att2}, it is found that the attention layer of baseline-NMT repeatedly attends input words ``newborn infant'' and ``30 ‐ year ‐ old or more''. 
Consequently, the decoder repeatedly outputs ``\begin{CJK}{UTF8}{min}新生児\end{CJK}'' (newborn infant) and ``\begin{CJK}{UTF8}{min}30歳以上の\end{CJK}'' (of 30 ‐ year ‐ old or more).
On the other hand, the attention layer of encoder-decoder-reconstructor almost correctly attends input words.

Table \ref{repete} shows a comparison of the number of word occurrences for each corpus and model.
The columns show (\romannumeral1) the number of words that appear more frequently than the counterparts in the reference, and (\romannumeral2) the number of words that appear more than once but are not included in the reference.
Note that these numbers do not include unknown words, so (\romannumeral3) shows the number of unknown words.
In all the cases, the number of occurrence of redundant words is reduced in encoder-decoder-reconstructor.
Thus, we confirmed that encoder-decoder-reconstructor achieves reduction of repeating and missing words while maintaining the quality of translation.

\section{Conclusion}
In this paper, we evaluated the encoder-decoder-reconstructor on English-Japanese and Japanese-English translation tasks.
In addition, we evaluate the effectiveness of pre-training by comparing it with a jointly-trained model of forward translation and back-translation.
Experimental results show that the encoder-decoder-reconstructor offers significant improvement in BLEU scores and alleviates the problem of repeating and missing words in the translation on English-Japanese translation task, and the encoder-decoder-reconstructor can not be trained well without pre-training, so it proves that we have to train the forward translation model in a manner similar to the conventional attention-based NMT as pre-training.

\bibliography{MyCollection}
\bibliographystyle{acl_natbib}

\end{document}